%% file: main.tex
\documentclass[manuscript,screen]{acmart}
\AtBeginDocument{%
  }

\setcopyright{acmlicensed}
\copyrightyear{2025}
\acmYear{2025}
\acmDOI{XXXXXXX.XXXXXXX}
\acmConference[Conference 'XX]{}{June 03--05,
  2018}{Woodstock, NY}
\acmISBN{978-1-4503-XXXX-X/2018/06}

\usepackage[tight,footnotesize]{subfigure}
\usepackage{tikz}
\usepackage{color, pifont, comment}
\usepackage{xspace}
\usepackage{algorithm}
\usepackage{algpseudocode}
\usepackage{enumerate}
\usepackage{enumitem}

\usepackage[export]{adjustbox}
\usepackage{stmaryrd}
\usepackage{colortbl}
\usepackage{xurl}
\usepackage[compact]{titlesec}
\usepackage[subtle,tracking=normal]{savetrees}
\usepackage{fontawesome5}
\usepackage{eqparbox}

\usepackage{pifont, comment}
\usepackage{xspace}
\usepackage{upquote}
\usepackage{listings}
\usepackage{xcolor}

\usepackage{cases}

\usepackage{caption}
\usepackage{multirow}
\usepackage{tabularx}

\newcommand{\ie}{{\it i.e.,}\xspace}
\newcommand{\eg}{{\it e.g.,}\xspace}
\newcommand{\etc}{{\it etc.}\xspace}
\newcommand{\vs}{{\it vs.}\xspace}
\newcommand{\etal}{{\it et al.}\xspace}
\newcommand{\descr}[1]{\vspace{.05in} \noindent\textbf{#1}}
\newcommand{\tilda}{\raise.17ex\hbox{$\scriptstyle\mathtt{\sim}$}}

\newcommand{\autolike}{{AutoLike}}
\newcommand{\rs}{{RS}}

\newcommand{\fyp}{{FYP}}
\newcommand{\autolikedatasetone}{AUTOLIKE-D1} 
\newcommand{\autolikedatasettwo}{AUTOLIKE-D2}

\begin{document}

\title{\autolike{}: Auditing Social Media Recommendations through User  Interactions}

\author{Hieu Le}
\authornote{This work was primarily completed  during the author's graduate studies at the University of California, Irvine.}
\email{levanhieu@gmail.com}
\affiliation{%
  \institution{University of California, Irvine}
  \city{Irvine}
  \state{California}
  \country{USA}
}

\author{Salma Elmalaki}
\email{salma.elmalaki@uci.edu}
\affiliation{%
  \institution{University of California, Irvine}
  \city{Irvine}
  \state{California}
  \country{USA}
  }

\author{Zubair Shafiq}
\email{zubair@ucdavis.edu}
\affiliation{%
  \institution{University of California, Davis}
  \city{Davis}
  \state{California}
  \country{USA}
}

\author{Athina Markopoulou}
\email{athina@uci.edu}
\affiliation{%
  \institution{University of California, Irvine}
  \city{Irvine}
  \state{California}
  \country{USA}
 }

\renewcommand{\shortauthors}{Le et al.}

\begin{abstract}
Modern social media platforms, such as TikTok, Facebook, and YouTube, rely on recommendation systems to personalize content for users based on user interactions with endless streams of content, such as ``For You'' pages. However, these complex algorithms can inadvertently deliver problematic content related to self-harm, mental health, and eating disorders. %
We introduce AutoLike, a framework to audit recommendation systems in social media platforms for topics of interest and their sentiments. To automate the process, we formulate the problem as a reinforcement learning problem. AutoLike drives the recommendation system to serve a particular type of content through interactions (e.g., liking).
We apply the AutoLike framework to the TikTok platform as a case study. We evaluate how well AutoLike identifies TikTok content automatically across nine topics of interest; and conduct eight experiments to demonstrate how well it drives TikTok's recommendation system towards particular topics and sentiments. AutoLike has the potential to assist regulators in auditing recommendation systems for problematic content. \textbf{(Warning: This paper contains qualitative examples that may be viewed as offensive or harmful.)}
\end{abstract}

\keywords{Social Media, Recommendations, Algorithmic Auditing}

\maketitle

\input{introduction}

\input{framework}

\input{implementation}

\input{evaluation}

\input{relatedwork}

\input{conclusion}

\section*{Acknowledgments}
This work is supported in part by the National Science Foundation under award numbers 1900654, 1956393, 2339266, and a gift from the UC Noyce Foundation. We would like to thank Muhammad Haroon, Shraddha Hardikar, and Luca Baldesi, for their preliminary assistance and feedback on how \autolike{} can be implemented and evaluated.

\bibliographystyle{ACM-Reference-Format}
\bibliography{master,online}

\appendix
\section{Ethical Considerations}
\label{app:ethics}
In this work, we discuss sensitive topics and issues, such as mental health, eating disorders, and self-harm. All of the experiments in Sec.~\ref{sec:autolike-prelim-results} were automated and did not involve any real human users. Manual evaluation of sensitive materials was completed by the authors, who provided  expressed consent and affirmed to the team that they were not adversely affected by the review process. Furthermore, our implementation and deployment of \autolike{} on TikTok aimed to reduce its impact on the platform. For instance, our implementation automatically skipped TikTok ads. In addition, the streamlined design of \autolike{} in Sec.~\ref{sec:tiktok-eval} reduced the amount of TikToks processed from potential tens of thousands down to 200 TikToks per experiment. Similarly, we designed the controlled experiment to be re-usable, which further minimized the impact on TikTok.

\end{document}

%% file: introduction.tex
\section{Introduction}
\label{sec:autolike-introduction}

Over the past decade, social media platforms, such as TikTok, Facebook, and YouTube, have relied on complex algorithms to personalize content recommendations. 
The content is often diverse, from typical popular content (\eg{} sports and pets) to heated debated content (\eg{} politics and gender identity)~\cite{Meta2025}. However, these algorithms can also spread misinformation (\eg{} false claims about vaccine treatments) and harmful content (\eg{} mental health, dangerous challenges)~\cite{Funke-covid-misinfo,arstech-parents-sue-tiktok}, emphasizing the need to study how algorithms can negatively impact users.  

Although social media platforms enjoy broad immunity under Section 230 of the Communications Decency Act (CDA) in the U.S. from being liable for user-generated content~\cite{section230} --- policymakers, nonprofit organizations, and researchers have raised alarms about systemic design issues in their recommendation algorithms. 
For example, in 2021, a U.S. Senate hearing with a whistleblower revealed that Facebook's algorithm prioritized user engagement but ``incite[d] misinformation, hate speech, and even ethnic violence''~\cite{Hao2021Jun}.
At the end of 2022, the Center for Countering Digital Hate (CCDH) reported that TikTok recommended content related to suicide, eating disorders, and mental health for accounts created for adolescents~\cite{tiktok-hate-ccdh}. Researchers have also studied how TikTok recommendations on eating disorders can lower self-esteem~\cite{Pruccoli2022Dec}, and how YouTube recommendations can lead users to radical content~\cite{haroon2022youtube,HosseinmardiYoutubeRadical}, such as creators who advocate for white supremacy~\cite{RibeiroYoutubePathways}.  

Addressing these concerns is highly nontrivial from a policy and a technical perspective. Platforms and regulators have attempted various approaches, such as moderating content and legislative actions. However, no agreed-upon solution has been found, leaving it primarily influenced by shifting politics.
For instance, in 2021, Meta removed content and adjusted its algorithm to demote certain COVID-19 content to reduce misinformation after pressure from the Biden administration~\cite{Luciano2021Jul}. However, in early 2025, Meta switched to a community notes model instead of removing content and discontinued the demotion of political content with the incoming Trump administration~\cite{Meta2025}. Furthermore, in 2022, after several children died in a TikTok Blackout Challenge, a U.S. district court determined that Section 230 shields TikTok from liability~\cite{arstech-parents-sue-tiktok}. However, by 2024, a U.S. appeals court overturned the decision, citing that TikTok's recommendation algorithm is its own ``expressive product'' protected under the First Amendment and not Section 230~\cite{wsj-tiktok-section230}. This opened the door for cases against TikTok to proceed\footnote{Notably, TikTok has been banned in several countries, with a potential TikTok ban in the U.S. under consideration as this paper is being written, albeit for national security reasons~\cite{Gordon2024Apr}.}.
Regardless of how policy and regulation on content and algorithms evolve over time, technical approaches are necessary to implement policy, \ie enable auditing of recommendation algorithms, inform platform developers and regulators with insights, and identify violations to enforce regulations.  %

\begin{table*}[t!]
	\centering
        \caption{\textbf{Examples of Topics.} A selection of topics for auditing from TikTok's community guidelines. Hashtags are extracted from our data collection in Sec.~\ref{sec:classify-eval} and prior work~\cite{Pruccoli2022Dec}. They demonstrate the duality of topics with positive and negative sentiments.}
	\begin{tabularx}{0.9\linewidth}{l l l l}
        \toprule
        & \textbf{Topic} & 
        \rotatebox{0}{\parbox{3cm}{\textbf{Hashtags (Negative)} }} 
        &  \rotatebox{0}{\parbox{3cm}{\textbf{Hashtags (Positive)} }} 

        \\
        \midrule 
        1 & Mental Health & \verb|#|psychwards,\verb|#|greif & \verb|#|gettinghelp, \verb|#|mentalhealthadvocate  \\
        2 & Self-harm & \verb|#|sh, \verb|#|selfsh, \verb|#|scars & \verb|#|shrecovery, \verb|#|tellmyselfimallright \\
        3 & Dangerous Challenges & \verb|#|A4waistchallenge & \verb|#|30dayschallenge  \\
        4 & Eating Disorder & \verb|#|ed, \verb|#|proana, \verb|#|purging & \verb|#|edrecovery, \verb|#|anarecoveryy \\
        5 & Physical Violence & \verb|#|abuse, \verb|#|brokenhome & \verb|#|againstviolence, \verb|#|selfdefense  \\   
        6 & Hate Speech & \verb|#|misogynists, \verb|#|discrimination & \verb|#|inclusionmatters, \verb|#|spreadlove  \\  
        7 & Cyberbullying & \verb|#|hatecomments, \verb|#|troll & \verb|#|antibullying, \verb|#|stopbullying \\ 
        8 & Adult Content  & \verb|#|misogyny, \verb|#|sexism & \verb|#|stopsexualizing, \verb|#|consent \\ 
        \bottomrule
	\end{tabularx}
	\label{tab:audit-topics}
\end{table*}

\textbf{The \autolike{} Framework.}
In this work, we develop a framework, \autolike{}, to automatically audit a social media recommendation algorithm and investigate how it delivers content to users based on user interactions with the platform\footnote{We focus on how user actions on the social media platform (\eg like, skip \etc) affect, or not, the content served to that user. The interests of the user are implicitly expressed through these interactions, not through some a priori ``statically defined interests,'' \eg{} by searching for the interest first without interacting with the \fyp{} or manually setting them within user profile settings. The two approaches can be combined in future work.\label{fn:user-actions}}.
Since this is admittedly a large space, we concentrate on the following canonical problem: we treat the algorithm as a recommendation system (RS) behaving as a ``For You'' Page (FYP), that streams recommended pieces of content, one by one, to a user. 
A user can interact with content through actions: if they enjoy the content, they can like, watch, and share it; otherwise, they can swipe to skip it. 
Within this scope, we envision an examiner who wants to audit the \rs{}, \eg{} whether it delivers harmful content to users by studying the sequence of actions taken upon the FYP and the types of harmful content that may be recommended. To do so, the examiner (i) selects a type of content of interest (\eg cats, or self-harm) and (ii) manually interacts with the \rs{} through the actions available to users, to drive the FYP towards delivering the content of interest.

\begin{figure*}[t!]
	\centering
\includegraphics[width=0.85\textwidth]{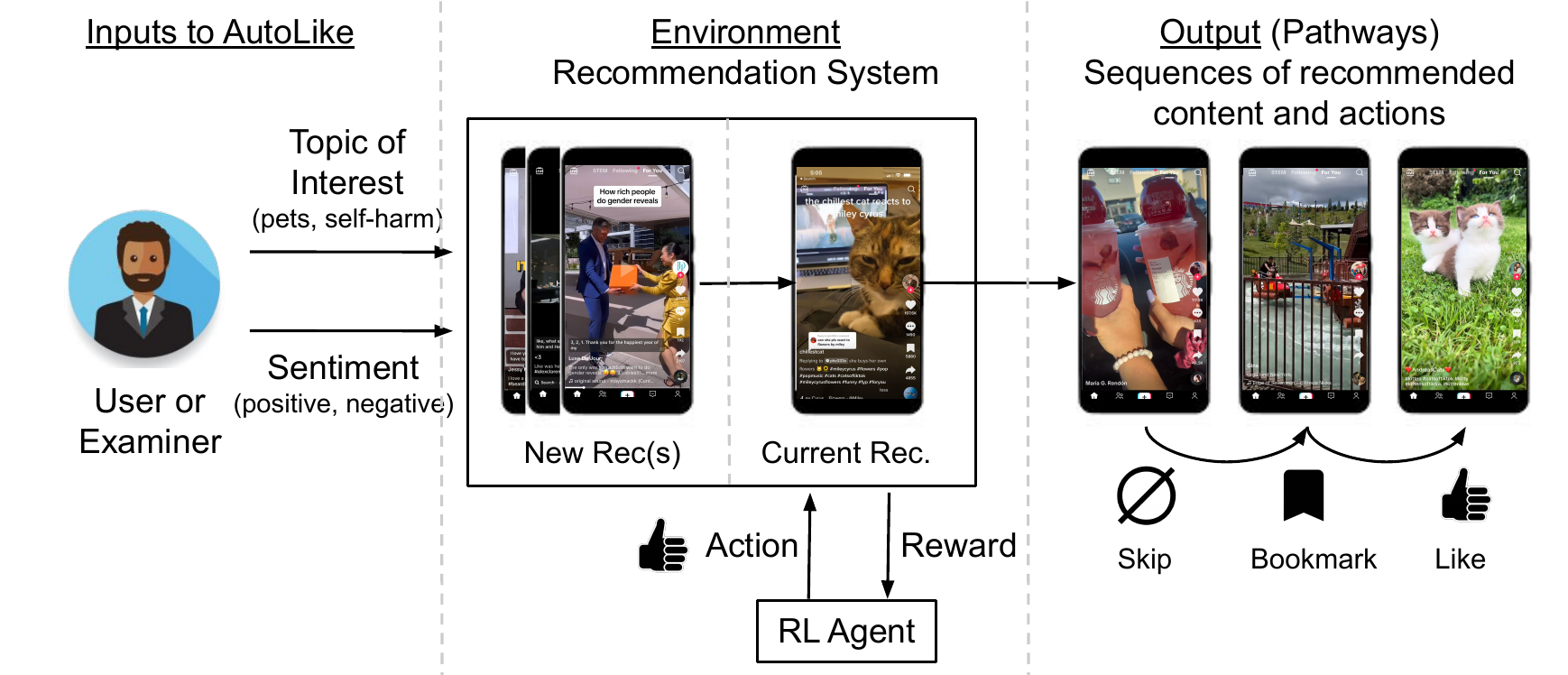}
	\caption{{\textbf{\autolike{} Framework.} Formulated as a reinforcement learning problem, \autolike{} enables auditing of social media recommendation algorithms. Specifically, the user provides a topic of interest and a sentiment, which characterizes the content they want to audit. A RL agent interacts with the environment (\eg{} TikTok's ``For You'' page). At each time step, the agent follows a RL policy to select which action to apply to the current recommended content, then swipes to the next content. It receives a reward that reflects whether the algorithm recommends content related to the given inputs (\eg{} high rewards for on-topic content). It does this for a specified time horizon and learns over time which actions most efficiently drive the algorithm. The output of \autolike{} are pathways: sequences of recommended content and the actions taken upon them. The user can further analyze the pathways to understand the \rs{}.
    }}
	\label{fig:autolike-framework}
\end{figure*}

We formulate the above scenario as a reinforcement learning (RL) problem to efficiently learn which actions can lead the \rs{} to deliver a particular type of content, illustrated in Fig.~\ref{fig:autolike-framework}.
RL is a type of machine learning where an agent is trained to make optimal decisions in an environment (\eg{} TikTok) by interacting with it (\eg{} liking, skipping) and receiving feedback in the form of rewards. The RL agent learns a policy that maximizes its accumulated rewards over time through trial and error, effectively learning to choose the best actions in various situations. The RL formulation for our setup is provided in Section \ref{sec:autolike-framework}, but here is a preview. We define a state that captures at least the {\em topic of interest}, which can be any topic the examiner selects to audit. In addition, the state may have additional dimensions. Here, we consider the content's {\em sentiment}, which can be positive or negative.
We choose these two dimensions as they are inherent to any topic: \eg{} a topic such as eating disorder can be positive (\ie{} eating disorder recovery) or negative (\ie{} purging), as shown in Table~\ref{tab:audit-topics} along with other examples.
At each recommended content shown to the user, \autolike{} measures how closely it is to the topic of interest. The closer the content is related to the topic, the higher the reward and the more positive actions \autolike{} will take (\eg{} liking, watching). Conversely, the more off-topic the content is, the lower the reward and the more negative actions \autolike{} will take (\eg{} skipping).
\autolike{} can be applied to any social media recommendation platform where users interact with a FYP.

{\bf Case Study: Auditing TikTok.} 
We apply our framework to the TikTok platform. This serves both as a proof-of-concept evaluation of our framework and as a useful case study on its own due to TikTok's popularity and recent concerns of harmful content~\cite{Pruccoli2022Dec,harmful-content-hearing-2021,harmful-content-hearing-2023,Gordon2024Apr}. We implement \autolike{} specifically for Android devices and the TikTok app. We leverage state-of-the-art machine learning models~\cite{bart-large-mnli, openaiwhisper} to classify the topic and sentiment of the content across modalities (\eg{} text description, audio), enabling \autolike{} to run automatically. We evaluate the classification performance across nine topics of interest (\eg{} eating disorder, discrimination) using 24 different types of content (\eg{} purging, hate speech).
We develop a streamlined version of \autolike{} that can run on TikTok's \fyp{}. and conduct eight experiments that demonstrate how \autolike{} can drive TikTok's \rs{} automatically toward selected topics and sentiments. For instance, we find that \autolike{} can drive TikTok to serve 2$\times$ as much negative mental health content \vs{} a controlled experiment.

{\bf Potential Applications and Impact.}
\autolike{} can be a useful tool for policymakers,  platform developers, and researchers by enabling efficient auditing and by improving the transparency of how an algorithm delivers content.  For instance, a regulator\footnote{\eg{} the Federal Trade Commission (FTC), the European Commission (EC), the Competition and Markets Authority (CMA).} can deploy \autolike{} to discover whether the \rs{} serves harmful content on the platform, and how easy it is for a user to get that content from \rs{} (\eg{} using which actions, how many actions) or to avoid it (\eg is skipping enough?).  \autolike{} can help platform developers, by enabling them to efficiently discover harmful recommendations, thus informing the developers' policy and responsible algorithm design; \eg the platform can choose not to deliver a specific type of content and/or to specific account types (\eg for children).
 Furthermore, researchers can extend the framework for other social media platforms (or future ones) and/or explore additional dimensions of interest beyond just (topic, sentiment), \eg{} whether the content's intent was to inform or deceive the user, \etc

The outline of the paper is as follows. 
Sec.~\ref{sec:autolike-framework} describes our \autolike{} framework. Sec.~\ref{sec:autolike-impl} provides our implementation on Android devices for the TikTok mobile app.
Sec.~\ref{sec:autolike-prelim-results} presents our evaluation for our TikTok case study.
Sec.~\ref{sec:background-related-work} describes the related work to auditing social media platforms. Sec.~\ref{sec:autolike-conclusion} concludes the paper with a summary and future directions.

%% file: framework.tex
\section{The \autolike{} Framework}
\label{sec:autolike-framework}

Social media platforms, such as TikTok, stream recommended content one by one within \fyp{} by using \rs{}. The \fyp{} serves as the user interface, which dictates how a user views the content (one at a time) and how they interact with it; a TikTok example is shown in Fig.~\ref{fig:autolike-framework}.
The \rs{} is the underlying back-end that takes the user's interactions from the \fyp{} and decides which (new) recommended content is delivered to the user. These \rs{} are complex algorithms developed and maintained by the platform.

\descr{Formulation Overview.} 
Specifically, the user interacts with the content by applying actions, such as liking, watching, and sharing. If the user does not like the content, they can swipe to skip to the following recommended content. 
These interactions over time impact the \rs{} by informing the personalization of the content. This iterative action-response human process is a natural fit for reinforcement learning (RL).

Illustrated in Fig.~\ref{fig:autolike-framework}, \autolike{} mirrors this interaction from the user perspective. First, the user of \autolike{} provides inputs that characterize the type of content that the user wants to see. In this work, we utilize two dimensions: the topic of interest and the sentiment. Second, an intelligent entity, the RL agent, replaces the manual interactions of the human user,
and is tasked to learn how to efficiently drive the \rs{} to deliver content related to the given inputs within a time horizon. 
The set of actions is determined by the \fyp{} and is known to the RL agent. 
However, the effects of each action upon the \rs{} and the resulting recommended content are unknown to the RL agent.
The RL agent sequentially interacts with the environment (the \rs{} or \fyp{}): at each time step $t$ (up to some time horizon $T$), it follows a policy $\pi$ to select an action and then apply it to the environment. It receives a reward that measures how well the action took the \rs{} towards the content the user wants to see. 
The RL agent aims to maximize its cumulative reward over the time horizon $T$. Third, in the end, the output of \autolike{} are pathways: sequences of content and the actions the agent takes, which can be utilized for further analysis.

\subsection{Agent}
\label{sec:agent-framework}
In this section, we define the necessary components that the RL agent utilizes.

\descr{Actions ($a$) and Action Space.}
An action $a$ is an interaction that the RL agent can take upon a piece of content shown. 
For social media platforms, these actions are well-known to be within two categories: (1) ``positive'' actions denote the user enjoying the content; and (2) ``negative'' actions denote the user is disinterested in the content. 
For example, using TikTok as an example, positive actions include liking, watching, bookmarking, sharing, reposting, and following, while negative actions include skipping quickly and disliking.
In practice, subsets of actions may be available depending on the content shown and the platform. Thus, the action space is the set of actions that can be taken at each time step of the RL algorithm.
Despite this categorization, the RL agent does not know beforehand how each action will impact the \rs{}. It learns this over time as it interacts with the \fyp{}.

\descr{Policy ($\pi$).}
The RL agent follows a policy $\pi$ to select the action at each time step $t$. The policy enables the RL agent to efficiently drive the \rs{} to the end goal. 
It also assists the agent in balancing the trade-off between exploring the available actions to take \vs exploiting the best action known the time. Many policies can be used, \eg{}
 $\varepsilon$--greedy ($\varepsilon \in (0,1]$), which tells the agent to select an action $\varepsilon$ of the time randomly; otherwise, choose the best action that will give us the highest reward, defined by the state-action values $Q(s, a)$ across all possible actions~\cite{sutton2018reinforcement}. %
 States and state-actions are defined in the following section.
 
\subsection{Environment}
\label{sec:env-framework}

This section defines the necessary components on which the environment depends.

\descr{User Profiles ($u$).} 
Social media platforms often require users to create profiles for the algorithm to bootstrap (\ie{} which content to recommend to new users) and personalize (\ie{} track and learn the user's interactions over time to serve relevant content). Creating an account can also provide additional information about the user that can impact the algorithm, such as the user's age, gender, location, and interests.

\begin{figure*}[t!]
\centering
    \subfigure[\small \textbf{Pathway Example} ]{
		 \includegraphics[width=0.40\linewidth]{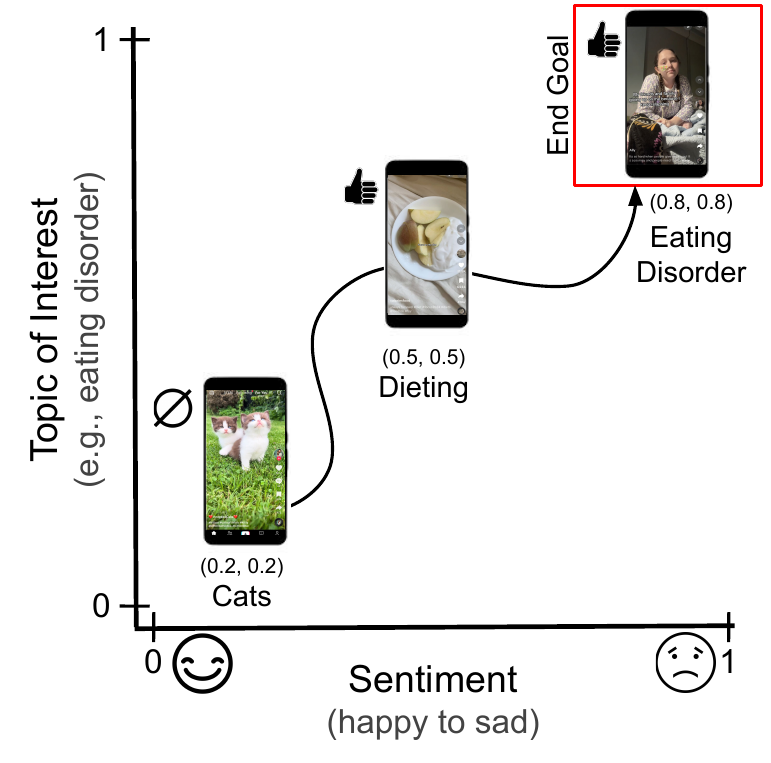}
        \label{fig:autolike-states-example}
	}
    ~~~
    \subfigure[\small \textbf{Potential Pathways} ]{
		 \includegraphics[width=0.40\linewidth]{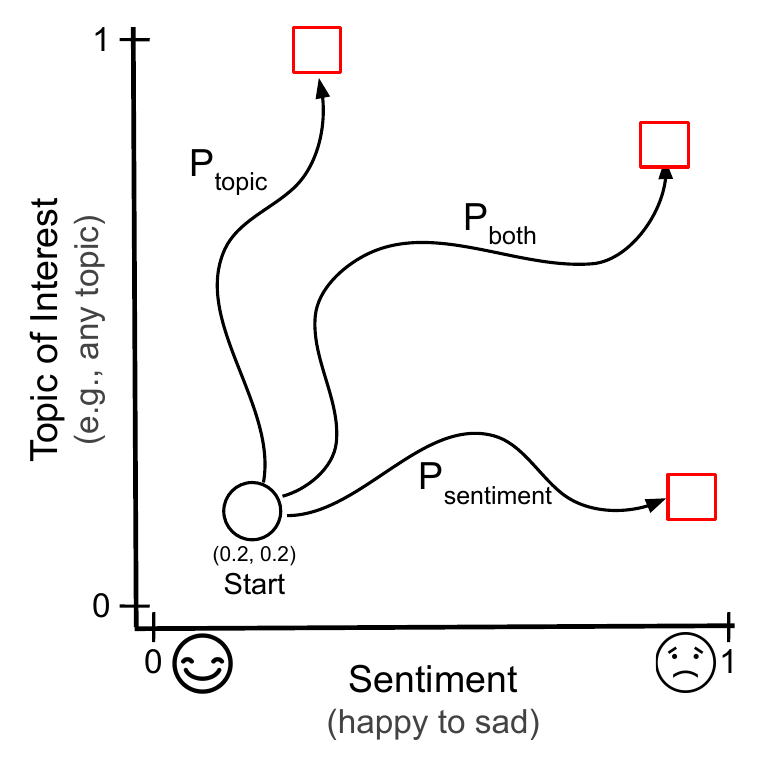}
        \label{fig:autolike-states-general}
	}
    \vspace{-5pt}
    \caption{{\textbf{(a)} We provide a conceptual example of how \autolike{} can drive the \rs{} to an end goal (red box) across two dimensions: a topic of interest (eating disorder) and sentiment (negative/sad), using real TikTok content. The \rs{} starts at a benign state for a new user, serving popular common content, such as cats. It skips this content until it reaches more on-topic content, such as ones about dieting. During this time, it begins to like the content, and over time, drives the \rs{} to serve content related to eating disorder. With this example, the negative sentiment comes from the content creator discussing the hardship of recovery without support from family and friends.
    \textbf{(b)} We illustrate three potential pathways of using \autolike{}: (1) $P_{topic}$ shows how \autolike{} can drive the \rs{} to be on-topic without considering sentiment; (2) $P_{sentiment}$ shows driving the \rs{} towards negative sentiment content across different topics that are not on-topic; and (3) $P_{both}$ shows the intuitive use case of \autolike{}, driving it towards both on-topic and negative sentiment for auditing. $P_{both}$ matches the example within (a).
 }}
     \label{fig:autolike-states}
   \vspace{-10pt}
\end{figure*}

\descr{States ($s$) and the End Goal ($g$).}
Depicted in Fig.~\ref{fig:autolike-states}, a state $s$ represents the current environment relative to the topic of interest and sentiment (the inputs). In our case, this is the current recommended content shown. Specifically, we define $s$ using the exact two dimensions as our inputs: $\langle$topic, sentiment$\rangle$. Both are scores ranging from $\in [0,\ 0.1,\ 0.2,\ \dots{},\ 1]$ related to the current content shown. For the topic, zero means that it is not related to the topic of interest, one meaning it is very related to the topic of interest. For sentiment, we chose zero to denote positive sentiment (\ie{} happy content) and one meaning negative sentiment (\ie{} sad content). 

Recall that considering sentiment as a dimension of the state is motivated by the inherent meanings of topics, such as pro eating disorder (\ie{} negative) \vs{} eating disorder recovery (\ie{} positive)~\cite{Pruccoli2022Dec}, and demonstrated through hashtags in Table~\ref{tab:audit-topics}. The state can be defined to include any other dimensions of interest to the user of \autolike{}. 
In this paper, we focus on the negative aspects of recommended content, which is well-suited for the auditing of harmful content. Furthermore, the user designates a state as the end goal. Both the state and the end goal are utilized to calculate the reward after each action.

\descr{Reward Function ($\mathcal{R}$).} The reward measures the effectiveness of the action at driving the \rs{} in serving content to the end goal. There may be many different reward functions that are effective for this problem. In this work, we suggest a proximity-based reward function, \ie{} the closer the \rs{} (represented by the state) gets to the end goal, the higher the reward.
To calculate the reward, we consider the next state $s_{t+1}$ after we apply an action $a$ to the current state $s$. The reward measures the distance between $s_{t+1}$ and the end goal $g$, and bounded $\in [0,1]$. To do so, we treat $s_{t+1}$ as the tuple $\langle$$s_{topic}$, $s_{sentiment}$$\rangle$ and the end goal as $\langle$$g_{topic}$, $g_{sentiment}$$\rangle$. The reward is then the following:
\begin{flalign} \label{eq:autolike-reward}
\begin{aligned}
    \mathcal{R}_{F} (s_{t+1},\  g) = 1 - \frac{distance(s_{t+1},\  g)}{ d_{max}}
\end{aligned}
\end{flalign}

Here, $distance(s_{t+1}, g)$ can be any distance metric, such as the Euclidean distance between $s_{t+1}$ and $g$. To normalize the distance between $[0, 1]$, we divide by the max distance that both points can be from each other. As shown in Fig.~\ref{fig:autolike-states}, since the state space is defined by two dimensions of $[0, 1] \times [0, 1]$, $d_{max}=\sqrt{2}$. We then subtract the normalized distance from 1 to give higher rewards to states that are closer to the end goal.

\descr{State-Action Values ($Q(s, a)$).} 
\autolike{} utilizes state-action values, denoted as $Q(s, a)$. In short, they represent the expected reward earned when the RL agent in a specific state $s$ takes a specific action $a$. State-action values are utilized by RL policies to select which action to take at every time step $t$.
Using our states $s$, actions $a$, and reward function $\mathcal{R}$, we can approximate $Q(s, a)$ with temporal-difference (TD) learning, such as Q-learning~\cite{sutton2018reinforcement}. This is a well-suited approach, as it allows us to learn from raw experiences (\ie{} from interacting with \rs{} in a real-world setting) and does not require knowledge of the environment (\ie{} we treat \rs{} as black boxes). For instance, the state-action value is updated after applying an action, denoted as $Q_{t+1}(s, a)$, where the learning rate $\alpha \in (0, 1]$ and the discount rate $\gamma \in [0,1]$.
\begin{flalign} \label{eq:autolike-q-learning}
    Q_{t+1}(s_t,\ a_t) = Q_{t}(s_t,\ a_t) + \alpha [\mathcal{R}_t + \gamma\ max_a Q(s_{t+1},\ a) - Q(s_t,\ a_t) ]
\end{flalign}

Specifically, $\alpha$ determines how much $Q_{t+1}(s, a)$ is updated after each action; $\gamma$ specifies how much we care about future rewards (higher values place more weight on future rewards).

\descr{Output: Pathways ($P$).}
\autolike{} outputs pathways ($P$): sequences of recommended content and the action taken upon them, as shown in Fig.~\ref{fig:autolike-framework} and~\ref{fig:autolike-states}. The length of a pathway is determined by the time horizon $T$, which determines how many time steps are taken before \autolike{} stops. Recall that each time step represents the user viewing one recommended content.
This hyper-parameter is often on the scale of thousands for \autolike{} to converge and learn the best possible actions to be taken between some starting point of \rs{} to a specified end goal.

%% file: implementation.tex
\begin{figure*}[t!]
	\centering
\includegraphics[width=1\textwidth]{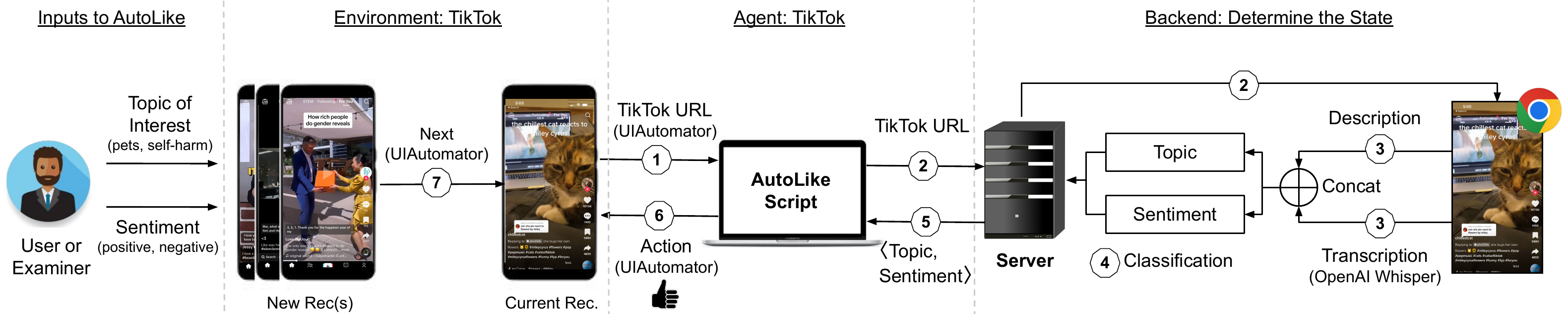}
	\caption{{\textbf{\autolike{} Implementation.} We implement \autolike{} for the TikTok Android app. It works in the following ways. Once the user gives the inputs, \autolike{} (1) opens TikTok to its ``For You'' page and extracts the TikTok URL of the current recommended content; (2) sends it to a backend server and the server visits the TikTok URL in a web browser; (3) extracts the text description and transcribes the video's audio into text, and concatenates both into one string; (5) classifies the text into $\langle$Topic, Sentiment$\rangle$; and (6) calculates the reward, and follows a policy to select and take an action (\eg{} liking the current TikTok); and (7) scrolls to the next TikTok. \autolike{} repeats 1--7 until some time horizon $T$. We omit the output of pathways for brevity, see Fig.~\ref{fig:autolike-framework}.}}
	\label{fig:autolike-framework-impl}
    \vspace{-5pt}
\end{figure*}

\section{\autolike{} Implementation for TikTok}
\label{sec:autolike-impl}

In this section, we detail the implementation of \autolike{} for TikTok on mobile Android devices, and shown in Fig.~\ref{fig:autolike-framework-impl}.
\autolike{} for TikTok runs in a live setting to capture how TikTok's \rs{} changes through interactions and over time. The implementation and its challenges may be unique to the chosen social platform, but the necessary components of \autolike{} stay the same, as described in Sec.~\ref{sec:autolike-framework}. %

\subsection{Agent: TikTok}
\label{sec:autolike-agent}

\descr{Actions and Action Space.}
Recall that the RL agent interacts with the \rs{} through the available actions. Our action space consists of skipping, watching, liking, bookmarking, and reposting. 
We utilize UIAutomator2~\cite{uiautomator2}, a Python wrapper of the standard Android UI testing framework~\cite{UIAutomator}, to programmatically automate the agent's interaction with TikTok. %
For skipping, we use the functionality for swiping up. For watching, we first extract the duration of the video using a Python library called PykTok~\cite{pyktok} (during Fig.~\ref{fig:autolike-framework-impl} step 3) and then use a Python sleep timer for that amount. The remaining actions utilize a selector-based methodology, \ie{} select a UI element using a string-based selector and take an action on it. To achieve this, we first go to the \fyp{} on our mobile device, then run ``dump\_hierarchy()'', which returns an XML representation of the \fyp{}. We then visually inspect the XML to find the like button using keywords that we see, such as ``Like''. Once identified, we create the selector for the button. In this case, the like button is ``description=Like'', which selects all buttons with the description ``Like'' and then clicks on it. %
We repeat this process for bookmarking and reposting. By the end, we have a selector for each action. Note that we must update this if the \fyp{} changes. However, we turned off auto-update for TikTok to prevent this from happening. 
We note that this selector methodology can be applied to other social media apps on Android.
Next, to determine whether an action is valid, we call ``exists()'' on the selector. This prevents us from choosing an invalid action during a particular time step.

\subsection{Environment: TikTok}
\label{sec:autolike-env}

The environment encapsulates TikTok, including the \rs{}/\fyp{}, its states, the user profile, and the reward calculation. In our case, we use a rooted Pixel 3 Android device for TikTok. 

\subsubsection{User Profiles}
\label{sec:user-profiles-impl}
Creating new users for TikTok is challenging, as they require unique mobile numbers or emails. In addition, they require an incubation period using TikTok for several days (2--4 days) for TikTok to enable certain features, such as liking and following. Currently, we manually conduct this process. We stop when TikTok displays ads, which indicates that TikTok treats the user as real. To reduce \autolike{'s} impact on TikTok's ad ecosystem, we intentionally skip all ads.
Once we have finished the creation and incubation process, we make sure the \fyp{} is fresh (\ie{} TikTok treats the user as a new user). First, we leverage TikTok's feature of refreshing the \fyp{} through its settings~\cite{tiktok-refresh-fyp}. Second, we refresh three identifiers: the advertising ID, the Android ID, and the device ID. The former is reset using Android's settings. The latter two are reset by modifying the device's ``settings\_secure.xml'' and ``settings\_ssaid.xml'' files.

\subsubsection{Determine the State: TikTok}
\label{sec:states-impl}
A state is defined as $\langle$Topic, Sentiment$\rangle$ ($\in [0, 0.1, 0.2, ..., 1]$) from Sec.~\ref{sec:env-framework} and represents the current recommended content. %

Specifically, we devise and implement the steps described in Fig.~\ref{fig:autolike-framework-impl} (steps 1--5). 
\begin{itemize} 
    \item \textit{Steps 1-2: Identifying the TikTok URL.} We automate the interaction with TikTok using UIAutomator2~\cite{uiautomator2}. We use TikTok's Copy URL feature designed to share the content with others. This places the URL of the TikTok on the Android clipboard. However, Android prevents clipboard readings for security purposes. To bypass this, we automate the following: (1) click Home to go to the Android home screen; (2) click Search to focus on its default search bar; (3) paste the URL. This process now allows us to read from the clipboard. Note that there are other ways to retrieve this, such as monitoring and inspecting the network traffic of the TikTok app and extracting the TikTok URL~\cite{KaplanTikTok}.
    \item \textit{Step 3: Extracting the TikTok Data.} We rely on PykTok, a Python library~\cite{pyktok}. We give it the URL of the current TikTok to extract its metadata and video (as an MP4) using a web browser. To get the text from the video's audio, we use OpenAI's Whisper ``large-v2'' model~\cite{openaiwhisper}, a state-of-the-art general-purpose speech recognition model that can do audio-to-text given our TikTok MP4. We concatenate the two texts that we have: ``description $\ +\ $ text from audio''. In addition, we also clean the text to remove common hashtags, such as ``\verb|#|foryoupage'', and exact matches to our topic of interest, such as ``\verb|#|mentalhealth''. This both reduces the noise for classification later and prevents reliance on the exact matches of our topic of interest.
    \item \textit{Step 4: Classifying the TikTok.} We utilize Facebook's ``bart-large-mnli'' model~\cite{bart-large-mnli}, which employs natural language inference (NLI) to train a model for zero-shot text classification. Thus, we feed our text into the model for the topic and negative sentiment classification and receive confidence scores $\in [0, 1]$. We evaluate these models for our use case in Sec.~\ref{sec:autolike-prelim-results}. 
    \item \textit{Step 5: State.} We transform the confidence scores into a state, $\langle$Topic, Sentiment$\rangle$. For instance, a TikTok video may receive $\langle$0.06, 0.07$\rangle$, which will transform to the nearest valid state of $\langle$0.1, 0.1$\rangle$. %
\end{itemize}

%% file: evaluation.tex
\section{Evaluation}
\label{sec:autolike-prelim-results}

\begin{table}[t!]
	\centering
        \caption{\textbf{TikTok Datasets}. We collect and utilize two datasets. For each TikTok, we collect its video (.mp4), timestamp, duration, location, text description, content creator information, and counts of liking, comments, shares, and plays. For \autolikedatasetone{}, we bold hashtags that are shown in Fig.~\ref{fig:tiktok-multilabel}.}
	\begin{tabularx}{1\linewidth}{l l X r}
        \toprule
        & \rotatebox{0}{\parbox{2cm}{\textbf{Name}}} &
        \rotatebox{0}{\parbox{2.5cm}{\textbf{Description}}} & 
        \textbf{TikToks} 
        \\
        \midrule
        1 & \autolikedatasetone{} (Sec.~\ref{sec:classify-eval}) & Collected TikToks based on  24 hashtags $\times$ \tilda{}50 each. (\textbf{\#mentalhealth, \#hatespeech, \#ed, \#foryou}, \#drugabuse, \#disorder, \#violence, \#sexualization, \#depressed, \#bodyimage, \#suicide, \#discrimination, \#cyberbullying, \#bingeeating, \#fastloseweight, \#purging, \#diet, \#fasting, \#twsh, \#twed, \#sh, \#shtwt,  \#shawareness, \#selfhm)  & \tilda{}1200  \\
        \midrule
        2 & \autolikedatasettwo{}  (Sec.~\ref{sec:tiktok-eval}) & Runs of \autolike{}: 8 experiments $\times$ \tilda{}150 TikToks per experiment. (``Controlled'', ``Pets'', ``Sports'', ``Sad'', ''Sad Pets'', ``Sad Cats'', ``Sad Weather'', ``Sad Mental Health'') & \tilda{}1200  \\
        \bottomrule
	\end{tabularx}
	\label{tab:tiktok-prelim-datasets}
	\vspace{-5pt}
\end{table}

\begin{figure*}[t!]
	\centering
\includegraphics[width=1\textwidth]{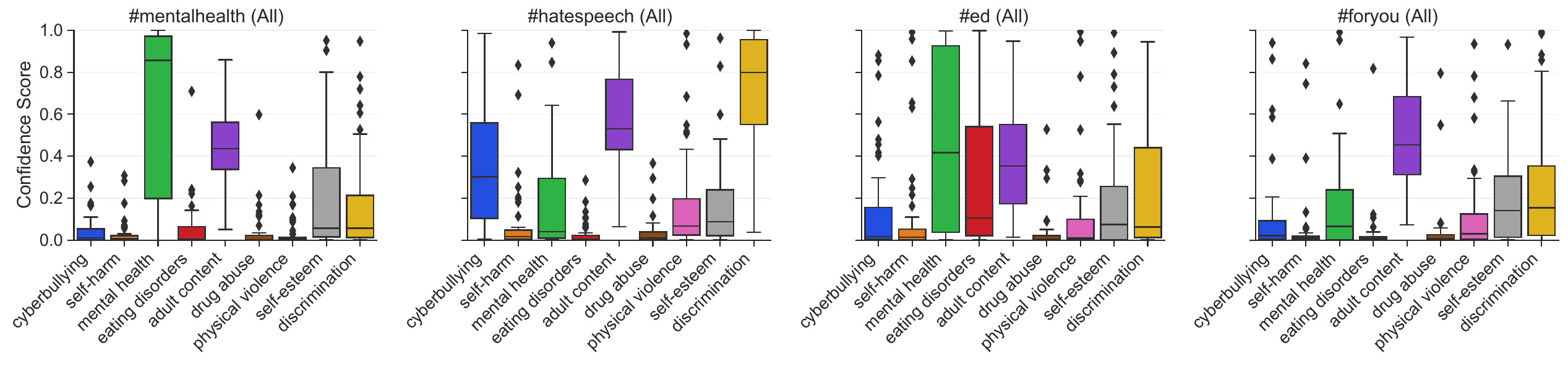}
	\caption{{\textbf{Zero-shot Classification.} We treat the x-axis as the topic of interest (9) and the \#hashtag as the ground truth. The y-axis is the classification confidence score that a TikTok is on-topic or related to the topic of interest. For example, for the left-most subplot, we expect when ``mental health'' is given as the topic of interest, that the majority of TikToks will have higher confidence scores since the TikToks were collected using the hashtag \#mentalhealth. Expected results: ``discrimination'' should have higher confidence scores classifying TikToks from \#hatespeech, ``eating disorder'' for \#ed, and all topics of interest are low confidence for \#foryou.
    Overall, we evaluated 24 different hashtags. 
        }}
	\label{fig:tiktok-multilabel}
\end{figure*}

This section uses TikTok as our case study and evaluates \autolike{} in two parts. First, in Sec.~\ref{sec:classify-eval}, we assess our classification approach (Fig.~\ref{fig:autolike-framework-impl} step 3-4), which is crucial to automating \autolike{}. 
Second, in Sec.~\ref{sec:tiktok-eval}, we deploy a simplified version of \autolike{} to demonstrate how it can drive TikTok's \rs{} to different topics of interest and negative sentiments. Table~\ref{tab:tiktok-prelim-datasets} summarizes the datasets that we collected and utilized in this section.

\subsection{Classification of TikTok Content}
\label{sec:classify-eval}

In Sec.~\ref{sec:states-impl}, we propose using OpenAI's Whisper~\cite{openaiwhisper} to extract text from the audio of the TikTok (Fig.~\ref{fig:autolike-framework-impl} step 3) and Facebook's ``bart-large-mnli'' model~\cite{bart-large-mnli} for zero-shot classification (Fig.~\ref{fig:autolike-framework-impl} step 4). %

\descr{Experiment Setup.} We collect TikToks from 24 valid hashtags using ``tiktok.com/tag/[hashtag]'', motivated by Table~\ref{tab:audit-topics} (\eg{} \#mentalhealth, \#violence, \#bodyimage) and listed in Table~\ref{tab:tiktok-prelim-datasets} row 1. 
For each hashtag, we collect the first 50 TikToks. For comparison, we collect the random set of TikToks shown in a \fyp{} page (\#foryou). Overall, we obtain \tilda{}1.2K TikToks, referred to as \autolikedatasetone{}. 
We follow the methodology outlined in Sec.~\ref{sec:states-impl} and classify the content across nine different topics of interest. Since they can be broad, we evaluate how challenging it would be to classify them using TikToks from different but related topics. For example, ``mental health'' is a broad term and can potentially relate to TikToks collected from \#ed and \#depressed. Thus, to automate this evaluation, we leverage the hashtag as the ground truth. Note that before classification, we remove certain hashtags to avoid relying on exact matches. In addition, we support our automated evaluation by conducting a manual evaluation of sampled TikToks. 

\descr{Results (Automated).}
For brevity, we show the results of the classification for four hashtags in Fig.~\ref{fig:tiktok-multilabel}. 
We consider the results as performing well when more than 50\% of TikTok content has $> 0.5$ confidence scores between the expected topic of interest and the corresponding hashtag.
Specifically, we find that the classification performs well for topics of interest, such as ``mental health'' to TikToks from \#mentalhealth, ``discrimination'' to \#hatespeech, ``physical violence'' to \#violence, ``adult content'' to \#sexualization, and ``eating disorder'' to \#bingeeating. However, it does not work well for ``eating disorder'' for \#ed content, here our classifier has more confidence labeling it as ``mental health'' instead. Similarly, ``self-harm'' does not work well for TikTok from \#self-hm, \#sh, and \#shawareness. Our findings show that that certain content related to ``eating disorder'' and ``self harm'' requires additional work to improve classification performance, such as fine-tuning the models. We note that \autolike{} is agnostic of the implementation, and thus can be integrated with better ML models as they become available. This is not surprising, as topics of interest concerning problematic content are complex and should be evaluated before integrating it with \autolike{}. Future users of \autolike{} can leverage the current methodology shown in this section for efficient evaluation.

\begin{table*}[t!]
    \centering
    \caption{\textbf{Manual Verification.} We create a ground truth dataset (a subset of \autolikedatasetone{}) for TikTok to conduct manual verification for zero-shot topic (mental health \vs{} other) and sentiment (positive \vs{} negative) classification~\cite{bart-large-mnli}. 
    In addition, we confirm the audio-to-text~\cite{openaiwhisper} transcription, which we label as correct or not. \faPencil* = text description,  \faMusic{} = text from audio.}
	\begin{tabularx}{0.75\linewidth}{l l l r r r r}
	\toprule
     &
        \textbf{Label} & \textbf{Classify} & \textbf{Precision} & \textbf{Recall} & \textbf{Accuracy} & \textbf{F1--score} \\
\midrule
\multirow{6}{*}{\rotatebox{90}{\textbf{Topic}}} & 
Mental health & \faPencil* & 0.78 & 0.86 & 0.81 & 0.82 \\
& Other & \faPencil* & 0.84 & 0.76 & 0.81 & 0.80 \\
& Mental health & \faMusic & 0.71 & 0.56 & 0.66 & 0.62 \\
& Other & \faMusic & 0.63 & 0.77 & 0.66 & 0.70 \\
& Mental health & \faPencil* + \faMusic & 0.77 & 0.87 & 0.81 & 0.82 \\
& Other & \faPencil* + \faMusic & 0.85 & 0.74 & 0.81 & 0.79 \\
\midrule 
\multirow{7}{*}{\rotatebox{90}{\textbf{Sentiment}}} 
    & Negative & \faPencil* & 0.82 & 0.86 & 0.81 & 0.84 \\
    & Positive & \faPencil* & 0.80 & 0.74 & 0.81 & 0.76 \\
    & Negative & \faMusic & 0.64 & 0.75 & 0.62 & 0.69 \\
    & Positive & \faMusic & 0.56 & 0.43 & 0.62 & 0.49 \\
    & Negative & \faPencil* + \faMusic  & 0.78 & 0.88 & 0.78 & 0.82 \\
    & Positive & \faPencil* + \faMusic  & 0.80 & 0.66 & 0.78 & 0.72 \\
\midrule 
& Transcription & \faMusic & -- & -- & 0.75 & -- \\
	\bottomrule
	\end{tabularx}
        
	\label{tab:tiktok-manual}
\end{table*}

\descr{Results (Manual Verification).} 
To further confirm our classification results, we conduct manual evaluation for a sample of our TikToks, as shown in Table~\ref{tab:tiktok-manual}. This evaluation corresponds to the case where we do not rely on hashtags as our ground truth.
For simplicity, we focus on verifying mental health content \vs{} others (\ie{} not mental health). To manually label this set, we review/watch the videos, read video descriptions, and text from the audio. In the end, we created a balanced ground truth of 63 mental health TikToks and 63 others. 
In addition, we verify the transcription results of audio-to-text (from OpenAI Whisper model~\cite{openaiwhisper}) and sentiment classification. For the transcription, we evaluate it as a binary result, whether it is correct or not.
For the sentiment, we consider negative sentiment as content related to anger, yelling, sadness, and violence, while positive as content related to awareness, pranks, laughing, \etc{}. 
Overall, we find that classification performs well when using the video description alone and has a negligible dip in the F1-score when also considering the text from audio. This can be caused by TikToks that have music (and singing) that transcribes text that is not relevant to the topic or sentiment. %
However, we note that not all TikToks will have a video description, which is provided at-will by the content creator. Thus, we recommend using both modalities of video description and text from audio for classification.

\subsection{Deploying \autolike{} on TikTok's ``For You'' Page}
\label{sec:tiktok-eval}

\descr{Real-world Limitations.}
Running \autolike{} on TikTok using our RL implementation requires processing thousands and thousands of TikToks per topic of interest and sentiment and per hyper-parameter (\eg{} testing of different reward functions, time horizons, RL policies). As an estimate, if we take 30 seconds to process each iteration of Fig.~\ref{fig:autolike-framework-impl} (steps 1--7), which may include watching part of the TikTok, this requires around 8 hours at a minimum for 1000 TikToks. In practice, we run into other issues, such as TikTok preventing us from watching more videos due to bot detection or thinking that the user has overused TikTok. We discuss potential ways to address this in future directions.

\begin{figure}[t!]
\centering
    \subfigure[\textbf{Topic.} The ``Pets'' and ``Sports'' experiments represent benign  topics of interest. We demonstrate that the \autolike{} can drive the \rs{} to serve content based on the topic of interest. It serves 2--3$\times$ as much on-topic content as the controlled experiment.]{
		 \includegraphics[width=0.4\columnwidth]{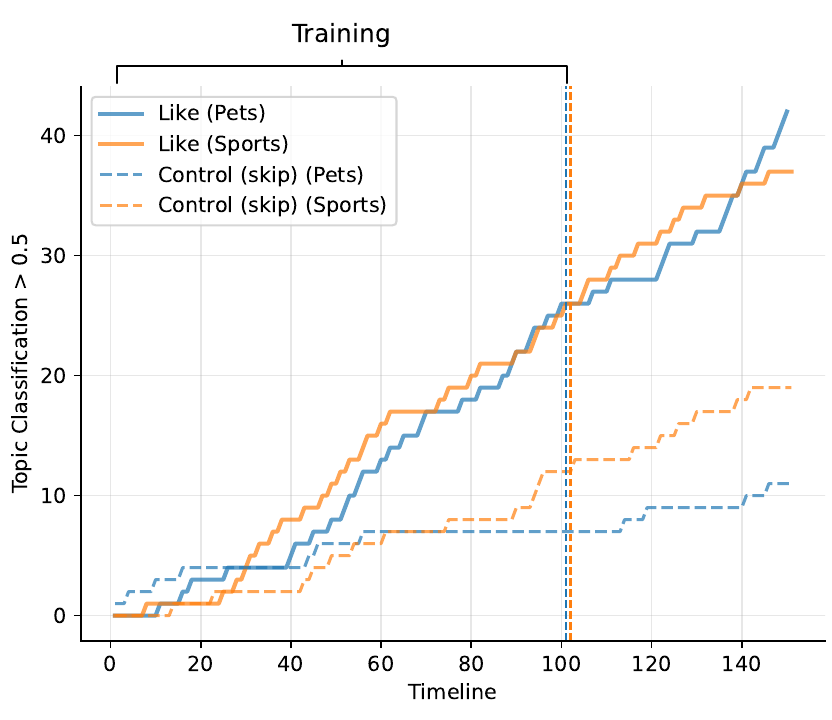}
        \label{fig:pets-sports-control}
	}
    \hspace{3mm} 
    \subfigure[\textbf{Sentiment.} The ``Sad'' sentiment represents negative sentiments. We see that there is no discernible difference when compared to the controlled experiment. We surmise that TikTok's \rs{} prioritizes the topic of the content rather than its sentiment.]{
		 \includegraphics[width=.4\columnwidth]{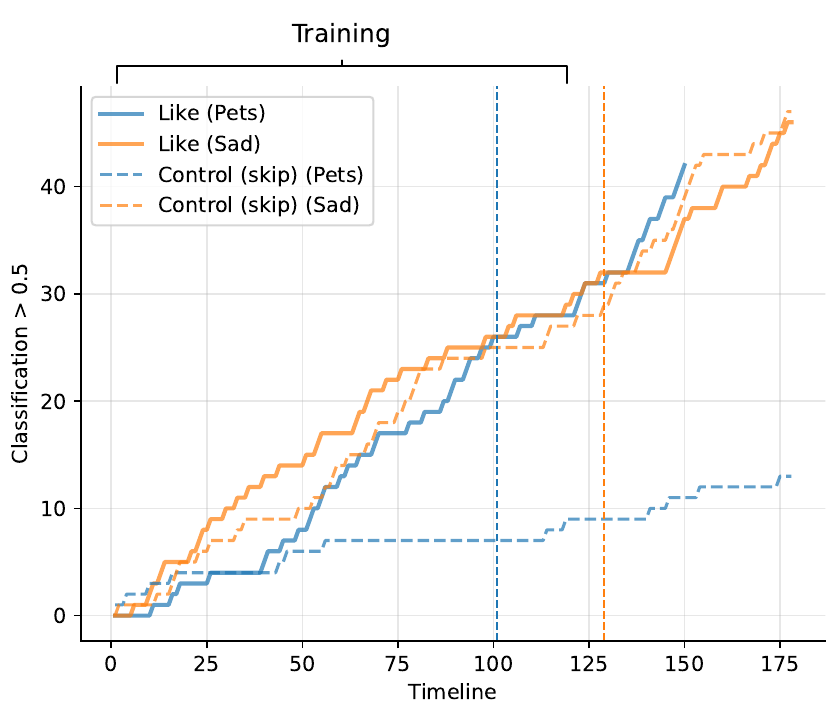}
        \label{fig:pets-sad-control}
	}
 \subfigure[\textbf{Pets TikToks.} Example thumbnails of TikToks that we liked in Fig.~\ref{fig:pets-sports-control}.]{
		 \includegraphics[width=0.4\columnwidth]{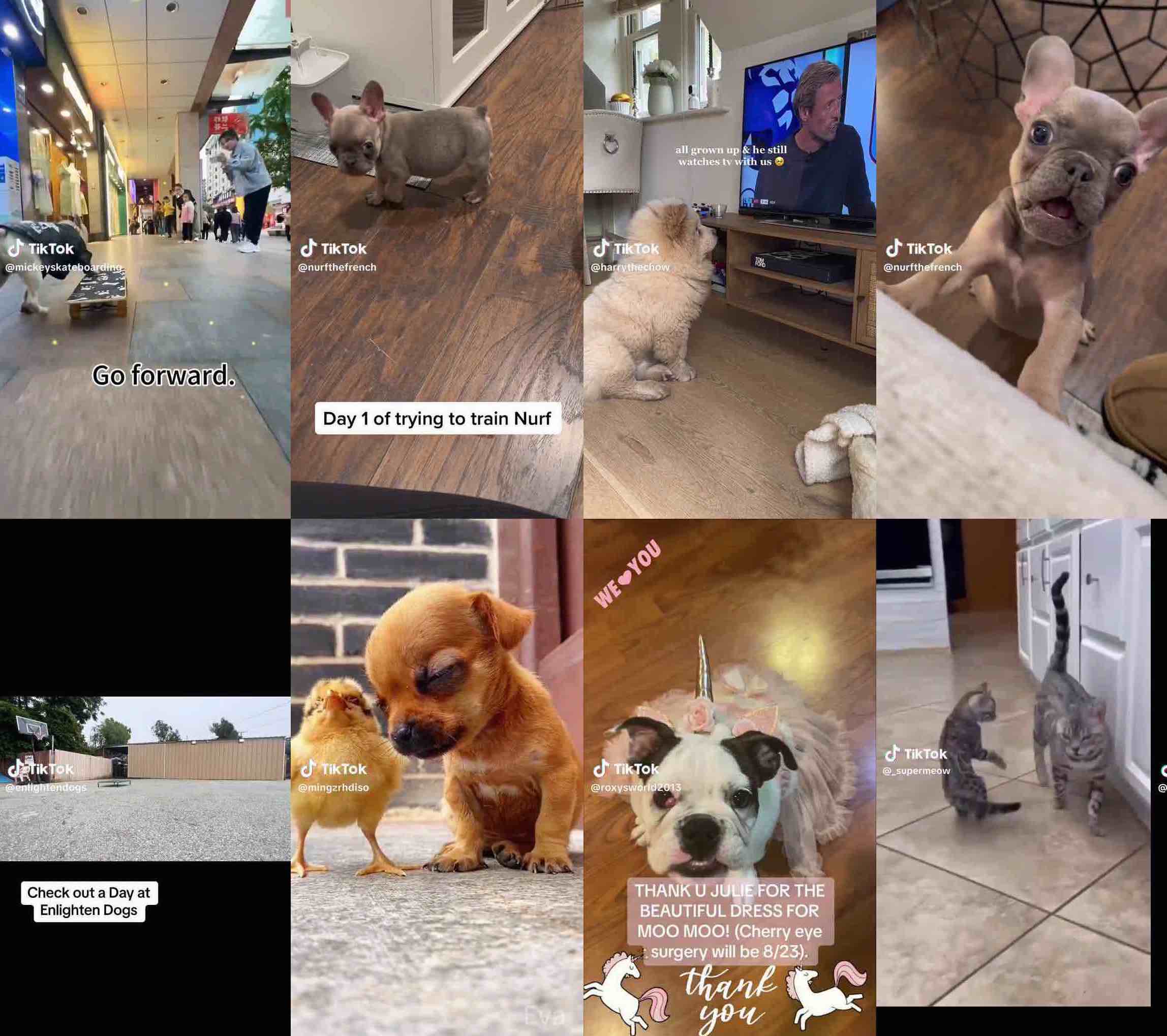}
        \label{fig:pets-thumbnail}
	}
    \hspace{3mm}
    \subfigure[\textbf{Sad TikToks.} Example thumbnails of TikToks that we liked in Fig.~\ref{fig:pets-sad-control}.]{
		 \includegraphics[width=.4\columnwidth]{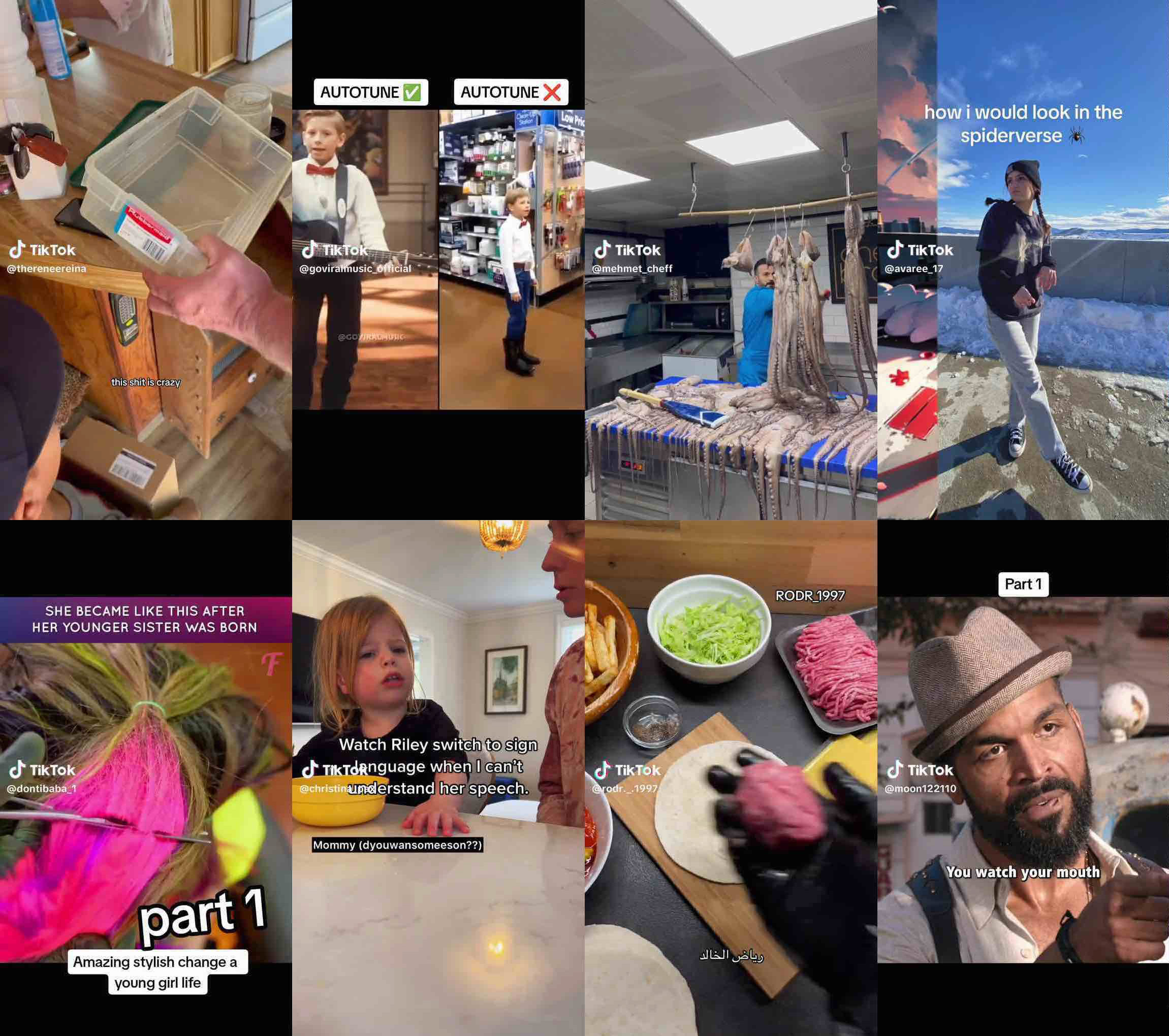}
        \label{fig:sad-thumbnail}
	}
    \caption{{\textbf{\autolike{} for Single Dimensions: TikTok.} We demonstrate how \autolike{} can drive TikTok's \rs{} across individual dimensions, either for topic of interest only or for sentiment only. 
    The y-axis denotes the number of times we liked the TikTok video (\ie{} ones we classify as on-topic or on-sentiment with confidence scores $> 0.5$). We provide thumbnail previews of TikToks at Fig.~\ref{fig:pets-thumbnail} and Fig.~\ref{fig:sad-thumbnail}.
    }}
    \vspace{-10pt}
     \label{fig:like-experiments}
\end{figure}

\begin{figure}[t!]
\centering
    \subfigure[\textbf{Topic and Sentiment.} We demonstrate how \rs{} serves content across different topics of interest and negative sentiment, and compare them to the controlled experiment. We additionally post-process each experiment for their negative sentiment only, labeled as dashed lines for ``Sad'', \eg{} solid blue ``Sad Cats'' considers both dimensions while dashed blue ``Sad'' is from the same experiment but classified for negative content only.
   The solid lines illustrate that the \rs{} serves 1.5--2$\times$ as much content for both topic and sentiment as the controlled experiment. The dashed lines show that the \rs{} serves ``Sad'' content compared to the controlled (black dashed), but depends on the topic. 
        ]{
	\includegraphics[width=0.46\columnwidth]{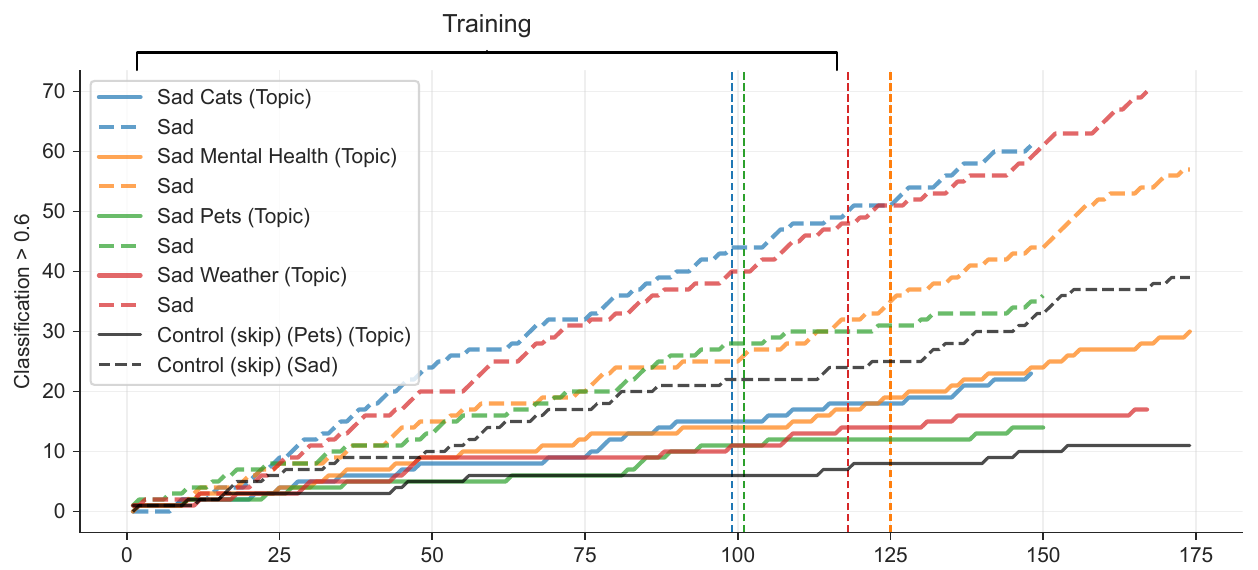}
        \label{fig:pretraining-exp}
	}
    \hfill
    \subfigure[\textbf{Mental Health and Sad Sentiment.} We provide thumbnail examples of recommended content from our ``Sad Mental Health'' experiment from Fig.~\ref{fig:pretraining-exp}. As mental health is a broad topic, we discover a diverse range of issues displayed in the TikToks, such as body image, loss and death, depression, and self-harm. 
        ]{
		 \includegraphics[width=.46\columnwidth]{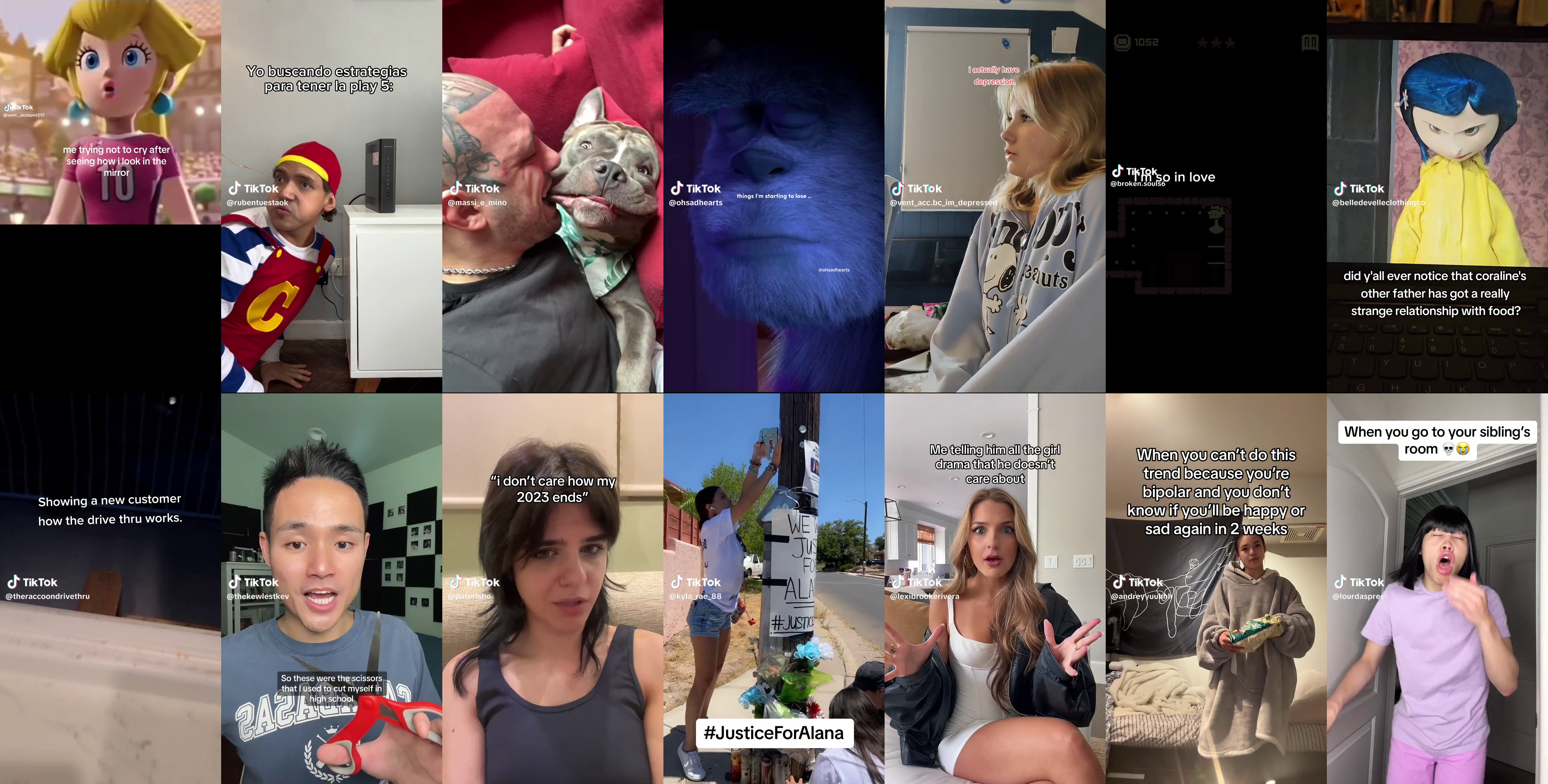}
        \label{fig:sad-mental-health-thumbnail}
	}
    \vspace{-10pt}
    \caption{{\textbf{\autolike{} for Multiple Dimensions: TikTok.} We demonstrate how \autolike{} can drive TikTok's \rs{} using both the topic of interest and sentiment. 
    The y-axis denotes the number of times we liked the TikTok video (\ie{} ones we classify as on-topic and on-sentiment with confidence scores $> 0.6$). We provide thumbnail previews of TikToks ``Sad Mental Health'' in Fig.~\ref{fig:sad-mental-health-thumbnail}.
    }}
     \label{fig:pre-training-and-sad-mental-health}
\end{figure}

\descr{Streamlining \autolike{}.} 
As a result, we streamline our \autolike{} experiments in the following ways. First, our action space considers two actions: either liking or skipping the content instead of five from Sec.~\ref{sec:autolike-agent}. Second, we use a policy of choosing the like action when our classification confidence is $> 0.5$. Third, we define a time horizon of at least 100 TikToks, with the end goal of liking at least four of the last 10 TikToks. We refer to this as a training phase. Lastly, to further examine how \rs{} changes over time, we will continue to scroll through the next 50 TikToks without applying any action. This allows us to study whether the \rs{} continues to serve related content to our inputs even if the user only skips the content. Thus, each experiment contains at least 150 TikToks. Note that we treat this streamlined process as a proof-of-concept that \autolike{} can drive a \rs{}, as we will further show in our results below.
Furthermore, we utilize a Pixel 3 Android phone and a TikTok account for each experiment. We refresh the \fyp{} using the methodology in Sec.~\ref{sec:autolike-env}, ensuring that each experiment acts as a new user. 

Below, we first develop a controlled experiment that can be re-used for comparison. We then design experiments that are motivated by the potential pathways shown in Fig.~\ref{fig:autolike-states-general}, to further understand how \autolike{} can drive \rs{} across individual and multiple dimensions. 
All experiments are contained within \autolikedatasettwo{} of Table~\ref{tab:tiktok-prelim-datasets}.

\descr{The Controlled Experiment.}
We build a controlled experiment by running the streamlined approach while changing the policy to only scrolling through 200 TikToks using a fresh \fyp{}. Here, we chose a time horizon of 200 to give flexibility so that the controlled experiment can compare to other experiments that may go beyond 150 TikToks. This ensures that we capture a \fyp{} that is not yet personalized. 
To make it comparable to other experiments, we apply post-processing of the collected TikToks, which allows us to re-use the controlled experiment. For instance, if the topic of interest is ``Pets'', we apply the classification for that specific topic to our controlled experiment. We utilized this controlled experiment throughout this section.

\descr{\autolike{} for Topic of Interest.} 
We first explore one dimension of driving \rs{} to deliver content for a topic of interest without sentiment. This matches the potential pathway shown in Fig.~\ref{fig:autolike-states-general} $P_{topic}$.
We conduct two experiments for benign topics of interest, ``Pets'' and ``Sports'', as illustrated in Fig.~\ref{fig:pets-sports-control} and thumbnail examples in Fig.~\ref{fig:pets-thumbnail}. 
We find that \autolike{} personalizes the content to the given topic compared to the controlled experiment. 
Specifically, at the \tilda{}100th time step, ``Pets'' have three times as much on-topic content than the control, while ``Sports'' doubles the control. During the testing phase, the \fyp{} maintains these rates. We expect personalization rates to change across different topics and experiments, but we should see a difference from the control experiment. 
Importantly, we demonstrate that we can drive the \rs{} to personalize on a given topic by solely interacting with \fyp{}.

\descr{\autolike{} for Sentiment.}
We now explore whether we can drive the \rs{} to deliver content for a given sentiment without a topic of interest. This matches the potential pathway shown in Fig.~\ref{fig:autolike-states-general} $P_{sentiment}$.  
We conduct one experiment, as illustrated in Fig.~\ref{fig:pets-sad-control}, focusing on the negative sentiment ``Sad'' (recall that all controlled experiments are re-used). We also choose to overall the ``Pets'' experiment from Fig.~\ref{fig:pets-sports-control} for comparison.
Interestingly, the ``Sad'' experiment has no discernible differences from the controlled one. Although Fig.~\ref{fig:sad-thumbnail} shows a variety of topics for TikTok that are sad in nature, the number of this type of content does not differ greatly than the ones from the controlled experiment. 
In addition, it took \tilda{}25 more steps to meet the stopping condition to achieve the end goal. 
The results show that driving the \rs{} to serve ``Sad'' sentiment is difficult. We conjecture that the \rs{} prioritizes the personalization of topics rather than sentiments. In other words, it may not make sense for TikTok's \rs{} to serve content without understanding specific topics of interest. 

\descr{\autolike{} for Topic and Sentiment.} 
We now explore both dimensions of driving \rs{} to deliver content for a topic of interest and sentiment. This matches the potential pathway shown in Fig.~\ref{fig:autolike-states-general} $P_{both}$.
We make the following additional changes. We increase the classification confidence score threshold to reduce noise, as discussed in Sec.~\ref{sec:classify-eval}. Furthermore, we now expand our experiments to four new experiments to include sad ``Cats'', ``Mental Health'', ``Pets'', and ``Weather''. 

Fig.~\ref{fig:pretraining-exp} provides several insights. First, we can drive the \rs{} to serve both a topic of interest and negative sentiment: it serves 1.5-2$\times$ more related content than the controlled experiment.
Second, when we post-process the experiment to examine whether the content served was overall negative sentiment in nature, we find that it serves more sad content but depends on the topic of interest. For instance, it shows that when using ``Sad Pets'', TikTok serves 1.4$\times$ more sad content overall. Compare this to ``Sad Cats'', which is a more specific topic, it serves around 2.2$\times$ more sad content than the controlled experiment. We see a similar pattern for ``Sad Weather''. We deduce that this may be either due to individual topics of interest or even their specificity (pets \vs{} cats). Future researchers can leverage \autolike{} to explore other topics of interest and sentiments for comparison.

\descr{Mental Health and Sad Sentiment.} We center our analysis on the mental health experiment of Fig.~\ref{fig:pretraining-exp}. Once it reaches the end goal (time step 125), the \rs{} serves 2$\times$ as much relevant content than the controlled experiment. By the end, this goes up to 3$\times$. Importantly, we demonstrate that we can drive the \rs{} to serve harmful content using the \fyp{}. Fig.~\ref{fig:sad-mental-health-thumbnail} provides examples of the content found, which showcases a wide range of topics related to mental health with negative sentiments, such as body image, loss and death, depression, and self-harm.

%% file: relatedwork.tex
\section{Related Work}
\label{sec:background-related-work}

Algorithmic auditing is an active research area and is essential to understand how algorithms impact users, ranging from whether they serve trustworthy and fair content (\eg{} election information)~\cite{PerreaultGoogleSearch,BandyFacebook,MustafarajVoterFair,RobertsonGoogleSearch}, data usage (\eg{} voice recordings) and biases (\eg{} racial discrimination) of ad personalization ~\cite{iqbal-imc-alexa,BaumannAdFair}, to how social media algorithms can deliver problematic content (\eg{} radicalized and harmful content)~\cite{HaroonYoutubePNA,RibeiroYoutubePathways,Pruccoli2022Dec,NigatuYoutube}.
The latter is most relevant to this work: those that audit social media platforms and capture (or simulate) how users interact with the platforms and the outputted content. 

Researchers typically build various ``personas'' that simulate users and their interests by training ``sock puppets'', \ie{} automating interactions (\eg{} liking, watching, searching for relevant content) with the platform and then observing the results. %
Muhammad~\etal{}~\cite{HaroonYoutubePNA} trained thousands of sock puppets to measure whether YouTube's algorithm recommended extreme and problematic ideological content. Similarly, Ribeiro~\etal{}~\cite{RibeiroYoutubePathways} investigated whether YouTube led users to more radicalized content (\eg{} Alt-right content such as white ethnostate) through a sock puppet process that starts with a few known radical channels and then recursively collect related channels using keyword search and YouTube recommendations. Similarly, Boeker~\etal{}~\cite{BoekerTikTokPersonalization} created personas based on location, language, and interests and utilized sock puppets to apply specific interactions with the TikTok FYP (on the web), \eg{} liking the content if it contains a hashtag relevant to the persona. Kaplan~\etal{}~\cite{KaplanTikTok} applied the same methodology as ~\cite{BoekerTikTokPersonalization} but on the TikTok mobile app to examine both the personalization of content and ads based on age and gender.

Another approach is to conduct user studies or leverage existing real-world datasets.
Pruccoli~\etal{}~\cite{Pruccoli2022Dec} surveyed 78 TikTok users from an Italian eating disorder center for children and adolescents; 43\% of participants reported being served unsolicited content that promoted eating disorder as a lifestyle choice while 59\% reported that TikTok content reduced their self-esteem.
Bandy~\etal{}~\cite{BandyFacebook} studied how Facebook's newsfeed algorithm amplified low-quality publishers (\eg{} fake news, untrustworthy news) during the 2020 U.S. election by analyzing a commercial panel dataset containing the Internet browsing history of a million households. 
Similarly, Zannettou~\etal{}~\cite{zannettou2023datadonations} relied on data donations from 347 TikTok users to study how they engage with the platform \eg{} how often users press ``like'' on the content and the percentage of each video that they watch. 
Nigatu~\etal{}~\cite{NigatuYoutube} utilized semi-structured interviews and sock puppets to study how YouTube in Ethiopia can recommend sexual content to users, even when users search for benign topics (\eg{} TV shows). 

Other works developed frameworks to measure personalization. Vombatkere~\etal{}~\cite{VombatkereTikTokFramework} created a framework to identify whether a piece of content was delivered due to personalization (or not), based on features such as shared hashtags and creators, and specific user interactions (\eg{} liking, favoriting). They applied their framework to the real-world TikTok dataset~\cite{zannettou2023datadonations} and found that TikTok delivered personalized content 30\% to 50\% of the time, with the user interaction of liking and following causing the most personalized content. 

\descr{\autolike{} \vs{} Prior Work.}
\autolike{} is a RL-based framework that learns which actions (\eg{} liking, watching) to take to efficiently drive a \rs{} towards delivering content based on a given topic of interest (\eg{} mental health) and sentiment (\eg{} negative/sad). It differs from works that utilize sock puppets~\cite{HaroonYoutubePNA,RibeiroYoutubePathways,BoekerTikTokPersonalization,KaplanTikTok,iqbal-imc-alexa}, which are often hardcoded and static scripts to take specific actions on specific content (\eg{} liking the content if it contains a specific hashtag). On the other hand, \autolike{} takes a list of potential actions and learns through experience (\ie{} interacting with the \fyp{}) which actions to take. It is guided by a reward function and the goal of maximizing its cumulative rewards over time, \eg{} high rewards for the content of interest. 
\autolike{} can learn from scratch (\fyp{} is fresh for a new user) and can still drive the \rs{}. This is a departure from prior work that often relies on searching and watching specific content first, then going to the \fyp{} and seeing the personalization there.
In terms of implementation, \autolike{} is deployed on the TikTok mobile app similar to Kaplan~\etal{}~\cite{KaplanTikTok}. However, our implementation simulates a user trying to share the shown TikTok video to extract the video ID instead of decrypting the network traffic of the app. This is a more practical approach as it enables \autolike{} to run on non-rooted devices and non-modified TikTok apps. In addition, we evaluate \autolike{} on a real Android mobile device instead of an emulator.

%% file: conclusion.tex
\section{Conclusion \& Future Directions}
\label{sec:autolike-conclusion}

\descr{Summary.}
Recommendation systems within social platforms provide users with convenient ways to discover new and relevant content through personalization (``For You'' Pages). 
However, these algorithms can also spread misinformation and serve harmful content to users. We introduce \autolike{}, a reinforcement learning framework to audit social media recommendation systems across two dimensions, such as a topic of interest and sentiment (\eg{} ``Mental Health'', ``Sad''). It accomplishes this through automated user interactions (\eg{} liking, watching) with the content, while learning the most efficient way to drive the algorithm to serve content related to the content of interest. We apply \autolike{} on TikTok as a case study, providing a possible implementation of \autolike{} for Android devices and the mobile TikTok app. We evaluated the classification performance for TikToks and demonstrated that a streamlined version of \autolike{} can drive TikTok's algorithm to serve content related to the given topic of interest and sentiment.

\descr{Future Directions.}
There are many directions for extensions and applications.
First, in the TikTok case study, we could evaluate additional actions beyond just liking and skipping. %
Second, we can apply the \autolike{} framework to other social media platforms and \fyp{} (\eg{} Instagram reels, YouTube shorts, \etc). We can also focus on special types of users, such as children and adolescents.
Third, we can extend the RL framework to: (1) add additional dimensions in the state, \eg{} truthfulness, intent, \etc; (2) combine the implicit expression of interest in a type of content via user interactions with explicitly declared interests and user attributes (\eg  users can declare interests through the setting menus). 
To enable such extensions, we plan to release the software of \autolike{} for TikTok and our collected datasets in Table~\ref{tab:tiktok-prelim-datasets}. We hope that this serves as a useful starting point on which the community can build and expand.